\pdfoutput=1

\documentclass[11pt]{article}

\usepackage[]{acl}

\usepackage{times}
\usepackage{latexsym}

\usepackage[T1]{fontenc}

\usepackage[utf8]{inputenc}

\usepackage{microtype}

\usepackage{booktabs}
\usepackage{multirow}
\usepackage{arydshln}
\usepackage{graphicx}
\usepackage{subcaption}

\usepackage{acronym}
\usepackage{etoolbox}
\makeatletter
\newif\if@in@acrolist
\AtBeginEnvironment{acronym}{\@in@acrolisttrue}
\newrobustcmd{\LU}[2]{\if@in@acrolist#1\else#2\fi}

\newcommand{\ACF}[1]{{\@in@acrolisttrue\acf{#1}}}
\makeatother

\acrodef{FT}[FT]{\LU{F}{f}ine \LU{T}{t}uning}
\acrodef{ICL}[ICL]{\LU{I}{i}n-cotext \LU{L}{l}earning}
\acrodef{PLM}[PLM]{Pre-trained Language Model}
\acrodef{MBE}[MBE]{Multilingual Bias Evaluation}
\acrodef{LLM}[LLM]{Large Language Model}
\acrodef{wrt}[w.r.t.]{with respect to}

\title{The Gaps between Pre-train and Downstream Settings in Bias Evaluation and Debiasing}

\author{Masahiro Kaneko$^{1}$ \quad
        Danushka Bollegala$^{2,3}$\Thanks{Danushka Bollegala holds concurrent appointments as a Professor at University of Liverpool and as an Amazon Scholar. This paper describes work performed at the University of Liverpool and is not associated with Amazon.} \quad
        Timothy Baldwin$^{1}$ \\
        $^1$MBZUAI \quad
        $^2$University of Liverpool \quad
        $^3$Amazon \\
        {\tt Masahiro.Kaneko@mbzuai.ac.ae} \\
        {\tt danushka@liverpool.ac.uk} \quad
        {\tt Timothy.Baldwin@mbzuai.ac.ae}
}

\begin{document}
\maketitle

\begin{abstract}

The output tendencies of \ac{PLM}s vary markedly before and after \ac{FT} due to the updates to the model parameters.
These divergences in output tendencies result in a gap in the social biases of \ac{PLM}s.
For example, there exits a low correlation between intrinsic bias scores of a \ac{PLM} and its extrinsic bias scores under \ac{FT}-based debiasing methods.
Additionally, applying \ac{FT}-based debiasing methods to a \ac{PLM} leads to a decline in performance in downstream tasks.
On the other hand, PLMs trained on large datasets can learn without parameter updates via \ac{ICL} using prompts.
\ac{ICL} induces smaller changes to \ac{PLM}s compared to \ac{FT}-based debiasing methods.
Therefore, we hypothesize that the gap observed in pre-trained and \ac{FT} models does not hold true for debiasing methods that use \ac{ICL}.
In this study, we demonstrate that \ac{ICL}-based debiasing methods show a higher correlation between intrinsic and extrinsic bias scores compared to \ac{FT}-based methods.
Moreover, the performance degradation due to debiasing is also lower in the \ac{ICL} case compared to that in the \ac{FT} case.

\end{abstract}

\section{Introduction}

\ac{PLM}s learn not only beneficial information~\cite{peters-etal-2018-deep,devlin-etal-2019-bert,brown2020language,touvron2023llama} but also undesirable social biases such as gender, race, and religous biases that exist in the training data~\cite{sun-etal-2019-mitigating,liang-etal-2020-towards,10.1162/tacl_a_00434,zhou-etal-2022-sense,guo-etal-2022-auto}.
Overall, two major approaches can be identified in the literature to elicit value from \ac{PLM}s in downstream tasks: \ac{FT} and \ac{ICL}.
\ac{FT} adapts \ac{PLM}s to specific tasks by updating parameters, while \ac{ICL} uses prompts without modifying the model parameters.

\begin{figure}[!t]
  \centering
  \begin{subfigure}{\linewidth}
    \includegraphics[width=\linewidth]{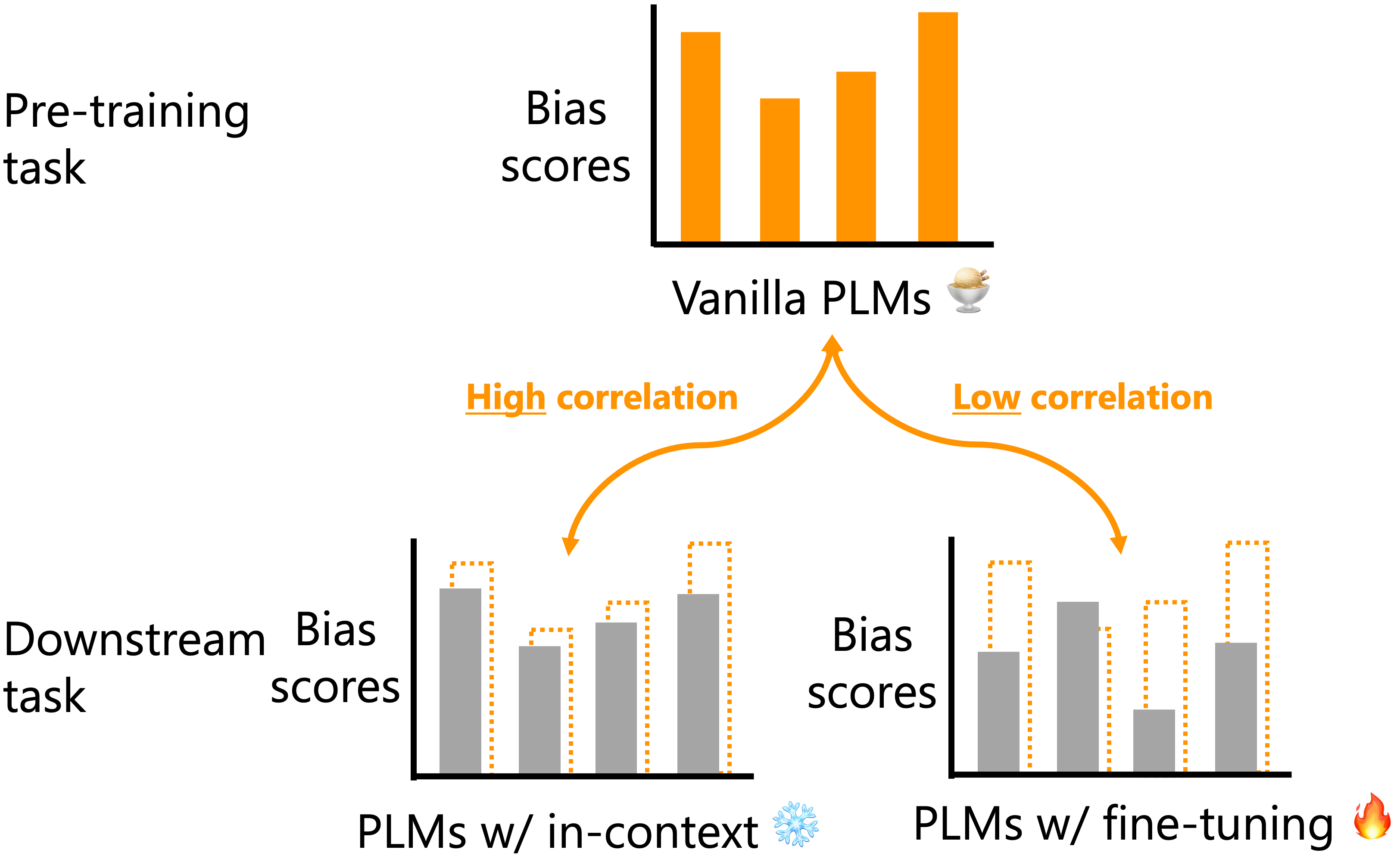}
    \caption{Bias evaluation.}
  \end{subfigure}
  \newline
  \begin{subfigure}{\linewidth}
    \includegraphics[width=\linewidth]{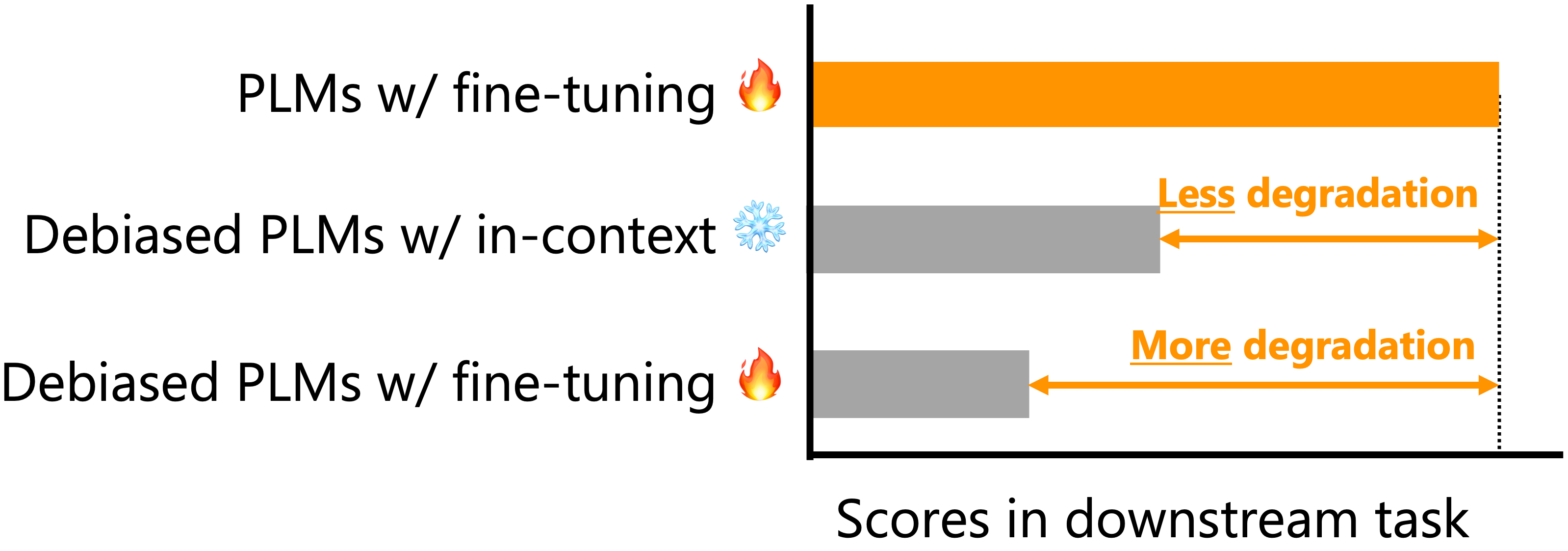}
    \caption{Debiasing.}
  \end{subfigure}
  \caption{The gap in bias scores when evaluating and debiasing \ac{PLM}s using \ac{FT}- and \ac{ICL}-based methods. A lower correlation between intrinsic and extrinsic bias scores (a), while a larger drop in downstream task performance (b) is encountered with \ac{FT} compared to \ac{ICL}.}
  \label{fig:abst}
\end{figure}

\ac{FT} models diverge considerably from the original \ac{PLM}s in their output distributions~\cite{chen-etal-2020-recall}.
Similarly, the output distribution of a \ac{PLM} is significantly affected by debiasing methods, because the parameters of the \ac{PLM} are updated during the debiasing process.
Debiasing accompanied by \ac{FT} suffers substantial performance decline in downstream tasks compared to the original \ac{PLM}~\cite{meade-etal-2022-empirical,kaneko2023impact,oba2023contextual}.
This is because the beneficial information learnt during pre-training is lost duringdebiasing.
Furthermore, bias evaluations exhibit a weak-level of correlation between pre-trained and \ac{FT} \ac{PLM}s~\cite{goldfarb-tarrant-etal-2021-intrinsic,kaneko-etal-2022-debiasing,cao-etal-2022-intrinsic}.

On the other hand, it is not obvious whether the prevalent wisdom regarding bias in such \ac{FT} regimes similarly pertains to \ac{ICL}, devoid of concomitant model updates.
The absence of parameter updates precludes the elimination of beneficial encodings, thereby minimizing adverse impacts on downstream task effectiveness.
\ac{ICL} strategies for mitigating biases may thus pose superior viability through obviation of representational damage.
Moreover, we hypothesize that bias evaluations contingent on the pre-training and downstream tasks exhibit heightened correlations, because the \ac{ICL}-based debiasing methods protect the model parameters.

In this paper, we investigate the performance gap of debiasing methods when applied to downstream tasks in an \ac{ICL} setting.
Additionally, we examine the correlation between bias evaluations for pre-training and downstream tasks enabled by the parameter sharing of \ac{ICL}.
Our experimental results show that \ac{ICL} has a smaller gap than the \ac{FT} setting \ac{wrt} performance degradation of debiasing and correlation between evaluations in pre-training and downstream tasks.
Therefore, we expect this paper to contribute by cautioning the community against directly applying trends from pre-training and downstream tasks with \ac{FT} to \ac{ICL} without careful considerations.

\section{Experiments}

We first explain the details of bias evaluations, debiasing methods, and downstream tasks used in our experiments.

\subsection{Bias Evaluations}

\paragraph{Pre-training settings.}
We target the following three intrinsic bias evaluation datasets.
\citet{nangia-etal-2020-crows} and \citet{nadeem-etal-2021-stereoset} proposed respectively, Crowds-Pairs (\textbf{CP}) and StereoSet (\textbf{SS}) bechmarks, which evaluate social biases of language models by comparing likelihoods of pro-stereotypical (e.g. \textit{``She is a nurse''}) and anti-stereotypical (e.g. \textit{``She is a doctor''}) examples.
\citet{kaneko-etal-2022-gender} introduced \ac{MBE} that evaluates gender bias in models in multiple languages by comparing likelihoods of feminine (e.g. \textit{``She is a nurse''}) and masculine (e.g. \textit{``He is a nurse''}) sentences.
Our research compares the bias scores in pre-training and the downstream tasks, which requires us to target the same language and bias type in both settings as considered in those benchmarks.
Therefore, we use gender bias in English on the above datasets to satisfy those requirements.

\paragraph{Downstream settings.}
We focus on three downstream tasks in our evaluations: question answering, natural language inference, and coreference resolution.
\citet{parrish-etal-2022-bbq} created the Bias Benchmark for Question answering (\textbf{BBQ}) to evaluate the social biases by determining whether a model predicts pro-stereotypical, anti-stereotypical, or unknown answers when given ambiguous and disambiguated contexts as the input.
\citet{anantaprayoon2023evaluating} proposed the Bias Natural Language Inference (\textbf{BNLI}) benchmark to evaluate gender bias using premise sentences and hypothesis sentences that share the same context but differ only in occupational and gender-related words.
\citet{zhao-etal-2018-gender} proposed the WinoBias (\textbf{WB}) benchmark to evaluate gender bias by considering differences between model predictions on pro-stereotypical and anti-stereotypical sentences in a coreference resolution task.

\subsection{Debiasing Methods }

\paragraph{Fine-tuning.}
We use the following two \ac{FT}-based debiasing methods that \textbf{modify the model parameters}.
\citet{webster2020measuring} introduced the Counterfactual Data Augmentation debiasing (\textbf{CDA}), swapping gender words in training data to reduce bias.
For instance, \textit{``She is a nurse''} becomes \textit{``He is a nurse''} in the augmented dataset, ensuring a more balanced model.
\citet{kaneko-bollegala-2021-debiasing} introduced All-Layer Token-level debiasing (\textbf{ALT}) for mitigating biases in MLMs.
It uses orthogonal projections to address gender and occupational terms, offering broad applicability and optimal performance with token-level debiasing across all layers of a \ac{PLM}.

\paragraph{In-context learning.}
We use the following two \ac{ICL}-based debiasing methods that \textbf{do not modify the model parameters} in our experiments.
\citet{ganguli2023capacity} presented a Zero-Shot Debiasing (\textbf{ZSD}) using the instruction -- \textit{``Please ensure that your answer is unbiased and does not rely on stereotypes.''} -- that is effective to mitigate social biases.
\citet{oba2023contextual} presented a Few-Shot debiasing \textbf{(FSD)} method in which examples are generated from manually designed templates representing counterfactual statements.
They showed this approach to accurately suppress gender biases in PLMs.

\subsection{Downstream Task Evaluations}

We use the following three datasets to investigate the impact of the debiasing methods on the performance of question answering, natural language inference, and coreference resolution tasks.
\textbf{RACE} dataset contains ca. 100K questions collected from the English proficiency
examinations for middle and high school students in China, covering a broad range of topics~\cite{lai-etal-2017-race}.
Adversarial Natural Language Inference (\textbf{ANLI}) dataset includes ca. 170K pairs and is collected via an iterative, adversarial human-and-model-in-the-loop procedure~\cite{nie-etal-2020-adversarial}.
\textbf{OntoNotes} v5.0 dataset has 13K sentences and is manually annotated with syntactic, semantic, and discourse information~\cite{pradhan-etal-2013-towards}.

\subsection{Pre-trained Language Models}

For the experiments, a \ac{PLM} needs to be of a size that allows efficient fine-tuning and be able to follow instructions for \ac{ICL}.
For this reason, we select the LaMini models~\cite{wu2023lamini} that are knowledge distilled from \ac{LLM}s using instruction data to create smaller models.
We used the following eight LaMini models\footnote{\url{https://huggingface.co/MBZUAI/LaMini-Neo-125M}}: LaMini-T5-61M, LaMini-T5-223M, LaMini-GPT-124M, LaMini-Cerebras-111M, LaMini-Cerebras-256M, LaMini-Flan-T5-77M, LaMini-Flan-T5-248M, and LaMini-Neo-125M.

We followed the same configuration as LaMini for fine-tuning, and used huggingface implementations for our experiments~\cite{wolf2019huggingface}.
We used four NVIDIA A100 GPUs for all experiments, and all training and inference steps were completed within 24 hours.

\subsection{Correlation between Bias Evaluations in Pre-training and Downstream Tasks}

\begin{table}[t]
\small
\centering
\begin{tabular}{lcccccc}
\toprule
& \multicolumn{3}{c}{Fine-tuning} & \multicolumn{3}{c}{In-context learning} \\
& BBQ & BNLI & WB & BBQ & BNLI & WB \\
\midrule
CP & 0.23 & 0.19 & 0.25 & 0.42 & 0.39 & 0.34 \\
SS & 0.20 & 0.15 & 0.20 & 0.38 & 0.44 & 0.42 \\
MBE & 0.10 & -0.02 & 0.12 & 0.29 & 0.35 & 0.31 \\
\bottomrule
\end{tabular}
\caption{Correlation between bias scores of intrinsic bias evaluation and extrinsic bias evaluation.}
\label{tbl:correlation}
\end{table}

In CP, SS, and MBE, each metric evaluates gender bias in the eight \ac{PLM}s mentioned above.
In BBQ, BNLI, and WB, we fine-tuned PLMs on downstream task datasets RACE, ANLI, and OntoNotes, respectively -- and evaluated gender bias \ac{wrt} bias evaluation in downstream tasks.
Furthermore, we used a few-shot \ac{ICL} setting where we provided the \ac{PLM}s with 16 randomly sampled instances from each downstream task dataset for FSD.
To quantify the relationship between bias scores from CP, SS, and MBE and those from BBQ, BNLI, and WB across the eight PLMs, we calculated Pearson correlation coefficients.
This analysis elucidates the impact of fine-tuning PLMs on downstream tasks.
Moreover, we show an evaluation of the original PLMs \ac{wrt} gender bias evaluations in pre-training and downstream tasks.

\autoref{tbl:correlation} shows the correlation between bias evaluation methods of pre-trained tasks (CP, SS, and MBE) and downstream tasks (BBQ, BNLI, and WB).
Overall, we see that \ac{FT} settings have low correlations between bias evaluations of pre-training and downstream tasks.
On the other hand, \ac{ICL} settings have higher correlations than \ac{FT} settings in every case.
Compared to \ac{FT}, \ac{ICL} has a relatively high correlation with bias evaluations in pre-training and downstream tasks, because it induces smaller changes to the model parameters.

Multiple existing work have reported a negligible correlation between pre-training and downstream task bias evaluation scores under the \ac{FT} setting~\cite{goldfarb-tarrant-etal-2021-intrinsic,cao-etal-2022-intrinsic,kaneko-etal-2022-debiasing}.
Currently, similar assumptions are applied to and discussed under \ac{ICL} settings as well~\cite{oba2023contextual,goldfarb-tarrant-etal-2023-prompt}.
However, \ac{ICL}-based debiasing results methods must be interpretted with special care.
Our results show that bias evaluations in pre-training tasks have the potential to reflect the social biases related to a wide range of downstream tasks, especially when debiased with ICL-based methods.

\subsection{Impact of Debiasing via Fine-tuning vs. \ac{ICL} in Downstream Task Performance}

\begin{figure}[!t]
  \centering
  \begin{subfigure}{\linewidth}
    \includegraphics[width=\linewidth]{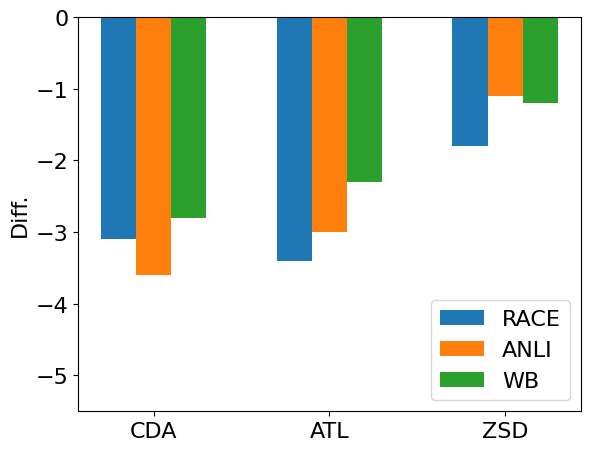}
    \caption{Bias mitigations are equalized w.r.t. ZSD.}
    \label{fig:diff-zsd}
  \end{subfigure}
  \newline
  \begin{subfigure}{\linewidth}
    \includegraphics[width=\linewidth]{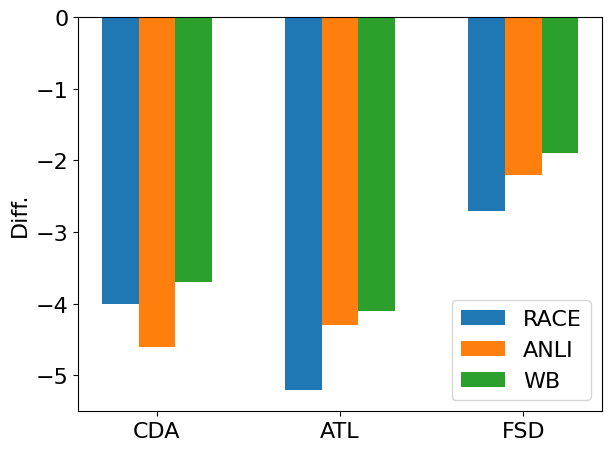}
    \caption{Bias mitigations are equalized w.r.t. FSD.}
    \label{fig:diff-fsd}
  \end{subfigure}
  \caption{Performance diffirence between original and debiased \ac{PLM}s in RACE, ANLI, and WB tasks are shown. Here, \ac{PLM}s are debiased using fine-tuning- (CDA, ATL) and \ac{ICL}-based methods.}
  \label{fig:diff}
\end{figure}

Debiasing methods decrease the downstream task performance of PLMs due to the loss of useful semantic information~\cite{kaneko-etal-2023-comparing}.
Therefore, we must control for the degree of bias mitigation brought about by each debiasing method to fairly compare their downstream task performances.
For this reason, we used a debiased model in which the debiasing results during the fine-tuning debiasing training fall within $\pm0.005$ of the debiasing score on the ZSD and FSD, respectively.\footnote{FSD is capable of adjusting the debiasing performance by varying the number of examples used.
In order to equalize the debiasing effects of FSD and ZSD, it would be necessary to reduce the number of FSD examples to 0.
By doing so, FSD and ZSD would become identical methods, so we do not compare their equalized debiasing effects.}

\autoref{fig:diff} shows the performance difference between the original and debiased models in RACE, ANLI, and WB tasks.
\autoref{fig:diff-zsd} and \autoref{fig:diff-fsd} show the effect of bias mitigation of CDA and ATL equalized respectively against ZSD and FSD.
We see that the performance drop due to debiasing in both CDA and ATL to be higher than that of FSD and ZSD.
Moreover, we see that the drop in performance of CDA and ATL to be higher when equalized \ac{wrt} ZSD than FSD, because ZSD imparts a lesser impact on the \ac{PLM} compared to FSD.
Overall, compared to debiasing via \ac{ICL}, debiasing via \ac{FT} results in a larger downstream task degeradation due to the updating of model parameters.

\subsection{Changing of Parameters in PLMs}

\begin{table}[t]
\small
\centering
\begin{tabular}{llcccc}
\toprule
& RACE & ANLI & OntoNotes \\
\midrule
CDA & 0.66 & 0.58 & 0.61 \\
ALT & 0.60 & 0.51 & 0.54 \\
\hdashline
ZSD & 0.81 & 0.83 & 0.87 \\
FSD & 0.73 & 0.76 & 0.81 \\
\bottomrule
\end{tabular}
\caption{Cosine similarity between output states of original and debiased models.}
\label{tbl:cos}
\end{table}

To quantify the change in model outputs due to \ac{FT} vs. \ac{ICL}, we measure the average similarity between the model outputs for a fixed set of inputs.
Specifically, we feed the $i$-th instance, $x_{i}$, from a downstream task dataset to the original (non-debiased) \ac{PLM} under investigation and retrieve its output state $e_{i}^{o}$ (i.e. the hidden state corresponding to the final token in the last layer).
Likewise, we retrieve the output states for the debiased model with \ac{FT} and \ac{ICL}, denoted respectively by $e_{i}^{f}$ and $e_{i}^{c}$.
We then calculate the cosine similarities ${\rm cossim}(e_{i}^{o},e_{i}^{f})$ and ${\rm cossim}(e_{i}^{o}, e_{i}^{c})$, and average them across the entire dataset as shown in 
\autoref{tbl:cos} for the eight LaMini \ac{PLM}s.
We can see that the cosine similarity is higher for the debiased models with \ac{ICL} than with \ac{FT}.
Therefore, debiased models with \ac{ICL} have smaller changes in output states than debiased models with \ac{FT}, indicating that the former is more likely to retain beneficial information from pre-training.
This result supports the hypothesis that the reduction of the gap in the relationship between pre-training and downstream settings is dependent on the changes in the parameters in the model due to debiasing.

\section{Conclusion}

We investigated the gap between pre-training and downstream settings in bias evaluation and debiasing and showed that this gap is higher for \ac{FT}-based debiasing methods than for the \ac{FT}-based ones.
Furthermore, we showed that the performance degradation in downstream tasks due to debiasing is lower in the \ac{ICL} settings than in the \ac{FT} setting.

Previous studies have referred to the results of \ac{FT} settings to discuss the relationship between pre-training and downstream settings~\cite{kaneko2019gender,kaneko-bollegala-2020-autoencoding,kaneko-bollegala-2021-dictionary,goldfarb-tarrant-etal-2021-intrinsic,cao-etal-2022-intrinsic}.
However, we emphasize that the settings of \ac{ICL} and fine-tuning differ in their tendencies and thus need to be discussed separately.

\section*{Limitations}

Our study has the following limitations.
We used the LaMini series~\cite{wu2023lamini} for our experiments because we needed to fine-tune models.
To investigate larger PLMs such as LLaMa~\cite{touvron2023llama} and Flan-T5~\cite{chung2022scaling} have the same tendencies, they need to be verified in environments with rich computation resources.
We only used QA, NLI, and coreference resolution as downstream tasks for our experiments.
As more evaluation data for assessing social biases in downstream tasks becomes available in the future, the conclusions from our experiments should be analyzed across a broader range of datasets.

There are numerous types of social biases, such as race and religion, encoded in PLMs~\cite{meade-etal-2022-empirical}, but we consider only gender bias in this work.
Moreover, we only focus on binary gender and plan consider non-binary gender in our future work~\cite{ovalle2023m}.
In addition, we only consider the English language, a morphologically limited language. As some research points out, social biases also exist in multilingual PLMs~\cite{kaneko-etal-2022-gender,levy-etal-2023-comparing}, which require further investigations.

\section*{Ethics Statement}

In this study, we have not created or released new bias evaluation data, nor have we released any models.
Therefore, to the best of our knowledge, there are no ethical issues present in terms of data collection, annotation or released models.
We observed that when employing ICL, there exists a correlation between intrinsic and downstream bias evaluations.
However, it must be emphasized that foregoing downstream bias evaluations and proceeding to deploy models presents a substantial risk.

\bibliography{custom}
\bibliographystyle{acl_natbib}

\end{document}